 \documentclass[letterpaper, 10 pt, conference]{ieeeconf}
\usepackage{graphicx}
\usepackage{float}
\usepackage{flushend} 
\usepackage{graphics} 
\usepackage{epsfig} 
\usepackage{mathptmx} 
\usepackage{times} 
\usepackage{amsmath} 
\usepackage{amssymb}  

\IEEEoverridecommandlockouts 

\title{\LARGE \bf Closed-loop underwater soft robotic foil shape control \\ using flexible e-skin}

\author{L.Micklem$^{1,4*}$\thanks{Corresponding author {\tt\small F.Giorgio-Serchi@ed.ac.uk}. Manuscript received: June, 19, 2024. $^{1}$Southampton Marine and Maritime Institute, University of Southampton, UK, $^{2}$Delft University of Technology, Netherlands, {$^{3}$Institute of Industrial Science, The University of
Tokyo, Japan}, {$^{4}$School of Engineering, The University of
Edinburgh, UK.} {$^{*}$Leo Micklem and Huazhi Dong contributed equally to this work.}}, H.Dong$^{4*}$, F.Giorgio-Serchi$^{4}$, Y.Yang$^{4}$, G.D.Weymouth$^{1,2}$, B.Thornton$^{1,3}$}
\date{}

\begin{document}

\maketitle

\begin{abstract} 

The use of soft robotics for real-world underwater applications is limited, even more than in terrestrial applications, by the ability to accurately measure and control the deformation of the soft materials in real time without the need for feedback from an external sensor. Real-time underwater shape estimation would allow for accurate closed-loop control of soft propulsors, enabling high-performance swimming and manoeuvring. We propose and demonstrate a method for closed-loop underwater soft robotic foil control based on a flexible capacitive e-skin and machine learning which does not necessitate feedback from an external sensor. The underwater e-skin is applied to a highly flexible foil undergoing deformations from 2\% to 9\% of its camber by means of soft hydraulic actuators. Accurate set point regulation of the camber is successfully tracked during sinusoidal and triangle actuation routines with an amplitude of 5\% peak-to-peak and 10-second period with a normalised RMS error of 0.11, and 2\% peak-to-peak amplitude with a period of 5 seconds with a normalised RMS error of 0.03. The tail tip deflection can be measured across a 30 mm (0.15 chords) range. These results pave the way for using e-skin technology for underwater soft robotic closed-loop control applications.   

\end{abstract}

\vspace{5mm}
\section{Introduction}

Soft robots are becoming prominent in a variety of fields \cite{Hoang2021,Drotman2021,Umedachi2016,Porez2014,Naclerio2021}. One area with potential for soft robots is bioinspired control surfaces and efficient actuation in subsea environments \cite{Renda2015,Aracri2021,Youssef2022}. Flexibility is important for efficient animal swimming and, by extension, swimming robots \cite{Quinn2022}. Flexible swimmers are known to exert fine-tuned control over their propulsor to excite different resonant modes and in this way maximize propulsive efficiency. Such a degree of control obviously necessitates a refined degree of authority over the morphology of the propulsors. This is currently inaccessible by soft robotics devices due to the lack of integrated underwater sensors with sufficient resolution for feedback control. 

Structural flexibility also allows for fine-tuning of the unsteady fluid loading on the propulsor with  benefits on sustained swimming efficiency as well as agile manoeuvring. For example, deformations which optimise the lift-to-drag ratio have been found to offset structural failure and hence facilitate survival in leaves subject to increasing wind speeds by maintaining low drag \cite{DeLangre2008}. This same hydrodynamic effect is optimised in swimming animals by adjusting their flapping amplitude and frequency according to their travelling speed. This complex control task is executed in fish via a combination of stiffness tuning through muscle contraction, fin and tail shape alteration, and skin surface changes \cite{Quinn2022}. In some fish, the control feedback required to exert authority on the foil morphology comes from fin rays acting as proprioceptive sensors \cite{Williams2013}. There is strong evidence that biological systems make use of shape feedback for efficient swimming and unsteady fluid load control. 

In their review of bioinspiration in underwater soft robotics, Youssef \textit{et al.} \cite{Youssef2022} highlight the need for all actuators and sensors to be implemented using completely soft materials. This is to not adversely impact the inherent flexibility of the system itself by incorporating stiffer components. Hegde \textit{et al.} \cite{doi:10.1021/acsnano.3c04089} divide the current soft robotic shape estimation sensors into three categories; resistive/piezoresistive, optical, and capacitive sensors. Resistive sensors work on the principle of varying resistance with pressure or strain. They are typically used to detect external pressures such as on a gripper but they struggle to perform shape estimation of highly non-linear soft robotic systems. Optical sensors use optical fibres to estimate shape and strain but are limited in the magnitude of deformation due to occlusion \cite{doi:10.1021/acsnano.3c04089}. 
Recently, Peng \textit{et al.} \cite{PENG2024} have used multiple inertial measurement units (IMUs) for shape estimation of soft manipulators. Reliance on IMUs and an Extended Kalman Filter offers a good degree of accuracy for the purpose of control but requires a reliable kinematic model of the body under investigation.
Hu \textit{et al.} \cite{hu2023} demonstrated a novel approach of utilising a specifically designed capacitive e-skin and deep learning which does not necessitate any pre-existing knowledge of the body and successfully demonstrated real-time, full 3D shape reconstruction of a flexible model robotic arm.  For shape feedback and reconstruction of underwater objects there are many methods using visual feedback \cite{Wang2022, Chadebecq2020, J2012, MIT1994}, thus suffering from the constraint of relying on cameras. However, there are currently no ready-made solutions for estimating the state of flexible underwater objects without the use of an external sensing system that does not disrupt the body's flexibility and that allows for use outside of a lab setting. This severely limits the ability of these methods to be used in real-world applications as in the case of an untethered soft underwater vehicles. 
 
Flexible foils have been studied extensively \cite{Alben2012} and have been shown to be good biomimetic models for understanding fish propulsion \cite{Shelton2014}. Rigid foils have also been widely used as control surfaces for Autonomous Underwater Vehicles (AUVs) \cite{DANTAS2013168}. These control surfaces are used for vehicle trajectory control and stability, and in some cases thrust generation \cite{BOWKER2022103148}. Flexible foils offer versatile options for soft underwater vehicle manoeuvring and propulsion,  but suffer of the lack of a suitable sensing technology for full state estimation. Therefore, shape control of a flexible foil is an essential stepping stone in the advancement of underwater soft robotics and  aquatic bioinspired locomotion.

\begin{figure}[t!]
    \centering
    \includegraphics[width=7.5cm]{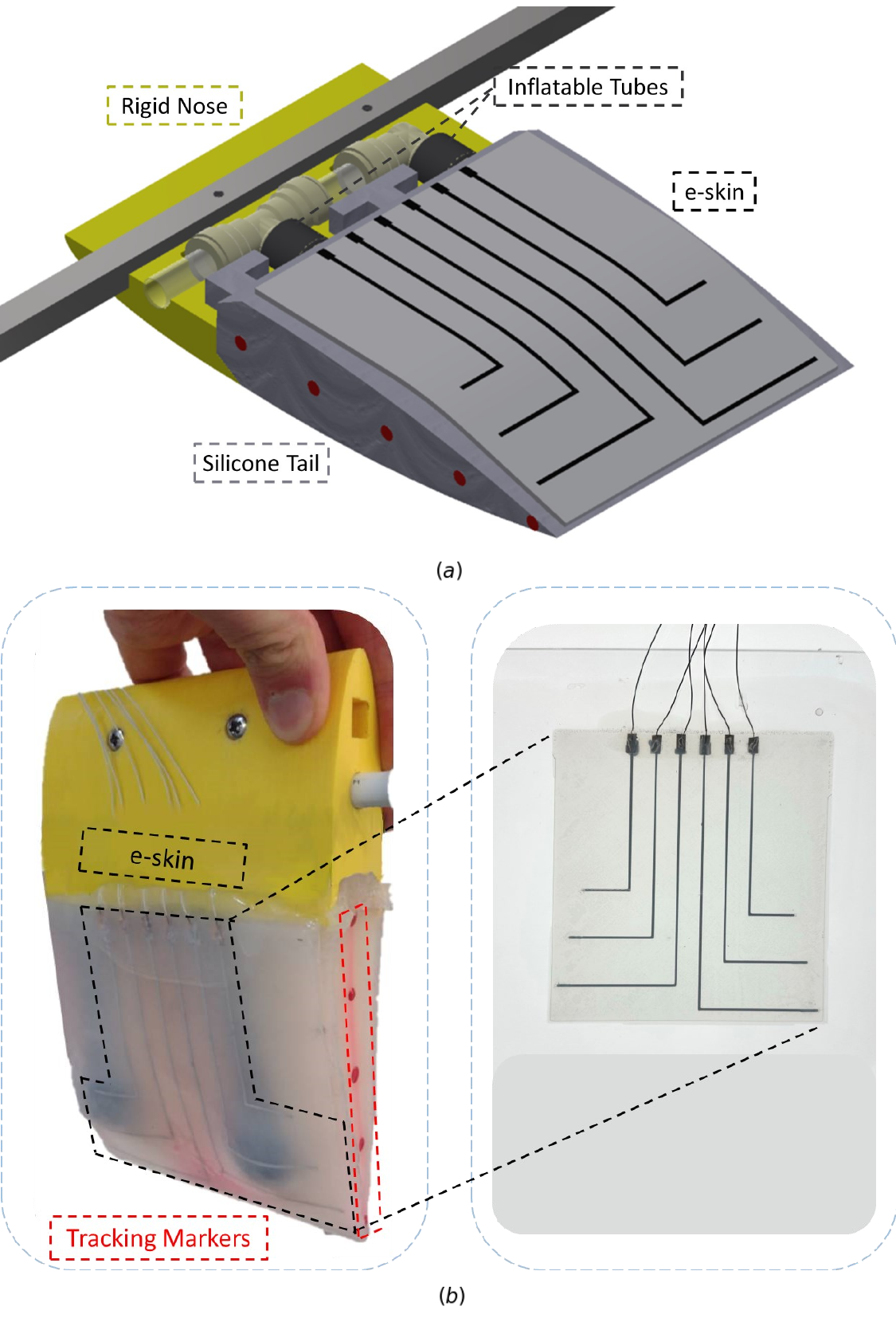}
    \caption{(a) Schematic of the tunable-stiffness soft foil. The rigid nose houses the internal pressure tubing, and clamps the silicone tail. The soft silicone tail has holes to house the inflatable rubber tubes which can expand and contract with pressure. The e-skin is bonded to the silicone tail using a thin layer of EcoFlex-30. (b) (Left) Soft robotic foil with e-skin attached for deformation measurement. The red tracking markers allow for the position of the foil to be tracked underwater for training and ground truth comparison. (Right) e-skin module for the soft robotic foil before attachment to the robot. 6 wires allow for the reading of 9 signals for training and measurement.}
    \label{fig:intro_plot}
\vspace{-0.3cm}    
\end{figure}

 
 In this work, we propose a 1 Degree of Freedom (DoF) test case of a flexible underwater foil with internal soft hydraulic actuators (Fig.~\ref{fig:intro_plot}a) \cite{Micklem2022}.  The actuators provide authority over the stiffness and camber with increasing pressure which results in the deformation of the foil in the form of increasing camber (Fig.~\ref{fig:camber explanation}). We develop the first working method of underwater shape control using a flexible capacitive e-skin sensor following \cite{hu2023}. We employ the flexible capacitive e-skin with machine learning to track the centre line of the flexible foil which allows the camber to be monitored as a control variable  (Fig.~\ref{fig:intro_plot}b). The camber is calculated in real time and fed to a PID controller to carry out a variety of time-varying set point regulation tasks. While the method is validated here on a foil undergoing planar camber control, the solution proposed is readily extendable to spatial deformations of complex multi-DoF aquatic propulsors. 

\section{Materials and Methods}
\subsection{Morphing Soft Foil}

We designed, built and tested a tunable-stiffness foil using our previously developed second-moment-of-area actuation \cite{Micklem2022} (Fig.~\ref{fig:intro_plot}a and Table~\ref{table_materials}). The foil is comprised of a rigid nose connected to a soft tail. Embedded within the tail are two inflatable elastic tubes. The tail has a base stiffness provided by the silicone, and the tubes can be pressurised to increase the second-moment-of-area and stiffness. When the tubes are inflated it causes the soft robotic foil to deform due to a natural curvature of the inflatable tubes \cite{Micklem2022}. The system is actuated hydraulically to avoid compressibility effects and negate buoyancy forces. 

\begin{figure}[t!]
    \centering
    \includegraphics[width=8.5cm]{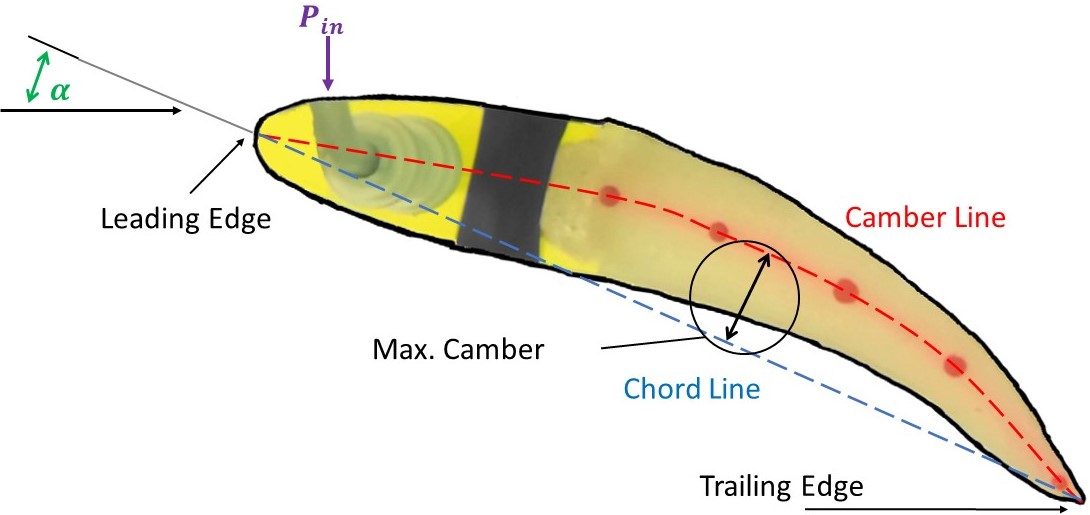}
    \caption{Outline of the key physical parameters for the control problem. The angle between any oncoming flow and the leading edge $\alpha$ is the angle of attack. The straight line from the leading edge to the trailing edge is the chord line. The line from leading edge to the trailing edge through the centre of the foil is the camber line. The perpendicular distance between the chord and the camber lines defines  the camber.}
    \label{fig:camber explanation}
\end{figure}

Hydraulic actuation affects the planar curvature of the foil, which is expressed here by the camber. The camber of a foil is described by its maximum value along the foil length as depicted in Fig.~\ref{fig:camber explanation} and expressed according to

\begin{equation}
    camber~\% = 100\times \frac{( Ch_n, C_n)_{max}}{Ch}
    \label{eqn:camber}
\end{equation}
where $Ch$ is the chord length, and $( Ch_n, C_n)_{max}$ is the maximum perpendicular distance between the chord line and the camber line, hence camber is given as a percentage. For a prescribed angle-of-attack $\alpha$ (Fig.~\ref{fig:camber explanation}), a change in camber of a foil results in changes in lift and drag forces. Typically, cambers are in the range of 0-10\%.  

\subsection{Design and Fabrication of the e-skin}

\begin{figure}[t!]
\centering
\includegraphics[trim= 0.cm 0.3cm 0.cm 0.cm, clip, width = 8.5cm]{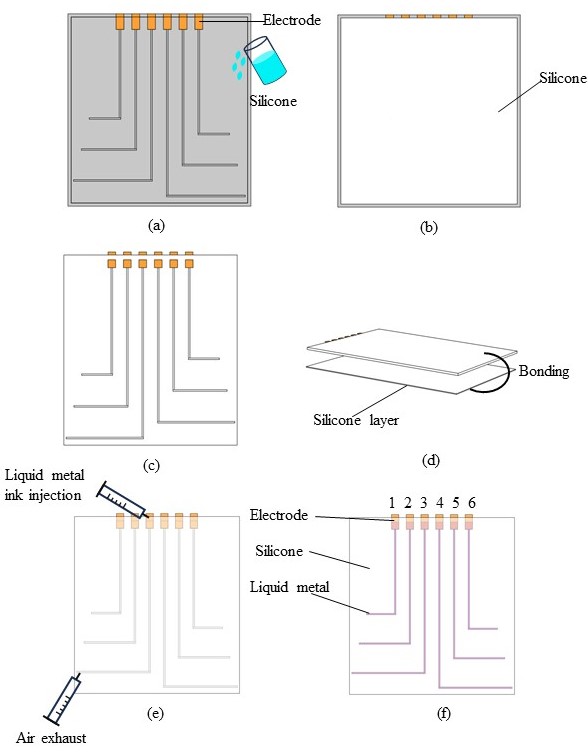}
\caption{Fabrication process of the capacitive e-skin. (a) Deployment of copper electrodes on the 3D-printed mould, (b) Eco-flex 00-30 is poured into the 3D printed mould, (c) curing of the top layer at room temperature for 4 hours and release the mould, (d) fabrication of an additional silicone backing layer and bonding with the top layer by means of the uncured silicone mixture as the adhesive, (e) injection of Liquid metal into the hollow channels with a second needle used as an exhaust for the air and finally sealing of the holes created by the needles with additional silicone. (f) The fabricated Liquid metal e-skin.}
\label{fig:e-skin manufacture}
\end{figure}

We developed a liquid metal-based e-skin for proprioceptive sensing of the flexible foil. Fig. \ref{fig:e-skin manufacture} shows the fabrication process and design of the capacitive e-skin. This consists of a silicone layer, 6 copper and liquid metal electrodes and a sealing layer. The overall dimensions are 120~$\times$~112 $\times$ 2~mm$^{3}$. The contact area of each electrode with the conductive layer is 5~$\times$~5~mm$^{2}$. A silicone elastomer, made of Ecoflex 00-30 (Smooth-On Inc.) is cast in a 3D printed mould to create vacant channels in for the liquid metal. The cured silicone is removed from the mould and bonded with a new silicone layer using uncured silicone mixture as the adhesive. Liquid metal is injected using a syringe into the hollow channels with a second needle used as an exhaust for the air. The holes created by the needles are then sealed with additional silicone. 

\begin{figure}[t!]
    \centering
    \includegraphics[width=0.49\textwidth]{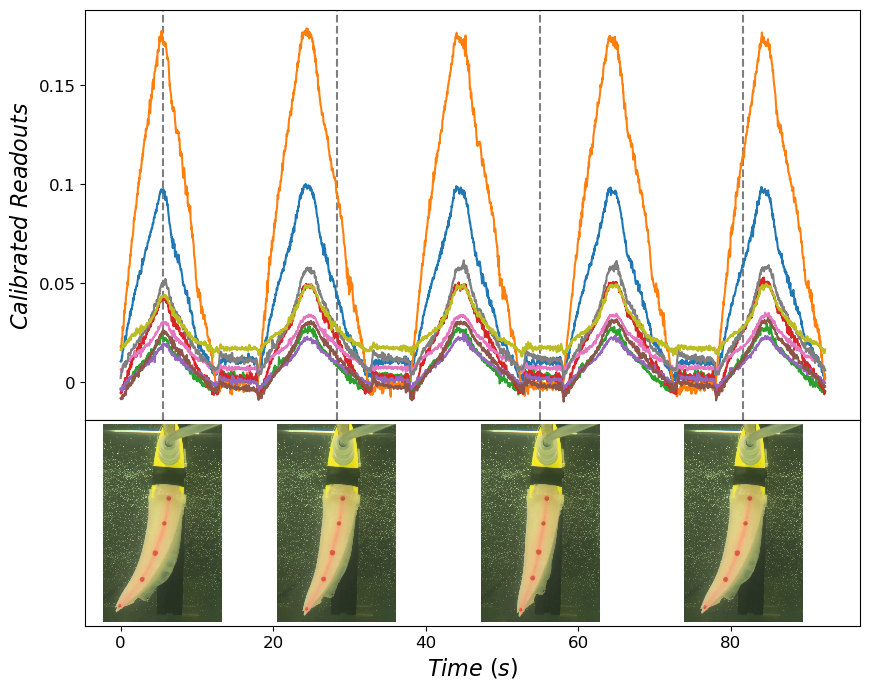}
    \caption{Signals of calibrated capacitance readouts, eq. \ref{eqn: skin calibration}, during foil deformation (top) and corresponding foil deformation (bottom) throughout five repetitive cycles of camber actuation.}
    \label{fig:Sensor Training}
\vspace{-0.3cm}
\end{figure}

The sensor operates by measuring the relative capacitance of select pairs of close proximity liquid metal electrodes with the strongest signals coming from adjacent pairs. As a result, we selected nine specific pairs of electrodes: (1, 2), (1, 3), (2, 3), (2, 4), (3, 4), (3, 5), (4, 5), (4, 6), and (5, 6), Fig. \ref{fig:e-skin manufacture}(f). When the shape of the e-skin changes due to deformation of the underlying foil, the relative capacitance between electrode pairs varies accordingly. By normalising the sensor readings against the undeformed shape readings, it is possible to use the capacitance signal to sense the foil's change in shape. Each readout is normalised as follows
\begin{equation}
    c = \frac{(c' - c_{emp})}{c_{emp}}
\label{eqn: skin calibration}
\end{equation}

 \noindent where $c$ is the normalised capacitance readout, $c'$ is the absolute readout, and $c_{emp}$ is the reference untrained readout. To quantify this change in shape, a model must be trained based on images of the robot taken at matching timestamps to the sensor readings, Fig.~\ref{fig:Sensor Training}. Upon successful training, reliance on the external visual feedback is no longer required. 

\begin{table}[b]
\caption{Experimental Materials}
\label{table_materials}
\begin{center}
\begin{tabular}{c c c }
\hline
 \multicolumn{3}{c}{Stiffness foil}\\
 \hline
 \emph{Part} & \emph{Material} & \emph{Dimensions (mm)}\\
 
Soft Tail & EcoFlex-30 & 120 x 140 x 30\\
Rigid Nose & Polylactic Acid Plastic & 120 x 80 x 30\\
Inflatable Tubes &  Isobutylene Isoprene Rubber & 110 x 15 x 15\\
Square Bar & Aluminium & 700 x 10 x 10\\
\hline

\multicolumn{3}{c}{Training Set Up}\\
\hline
DAQ & Bespoke Measurement Board &\\
Motor Driver & Cytron MD10C &\\
Controller & Arduino Uno&\\
Camera & GoPro HERO 10&\\
Resolution & 1920 x 1080&\\
Frame Rate & 30 fps&\\

\end{tabular}
\end{center}
\end{table}

\subsection{Underwater E-skin Training}

\begin{figure}[t!]
    \centering
    \includegraphics[width = 8.5cm]{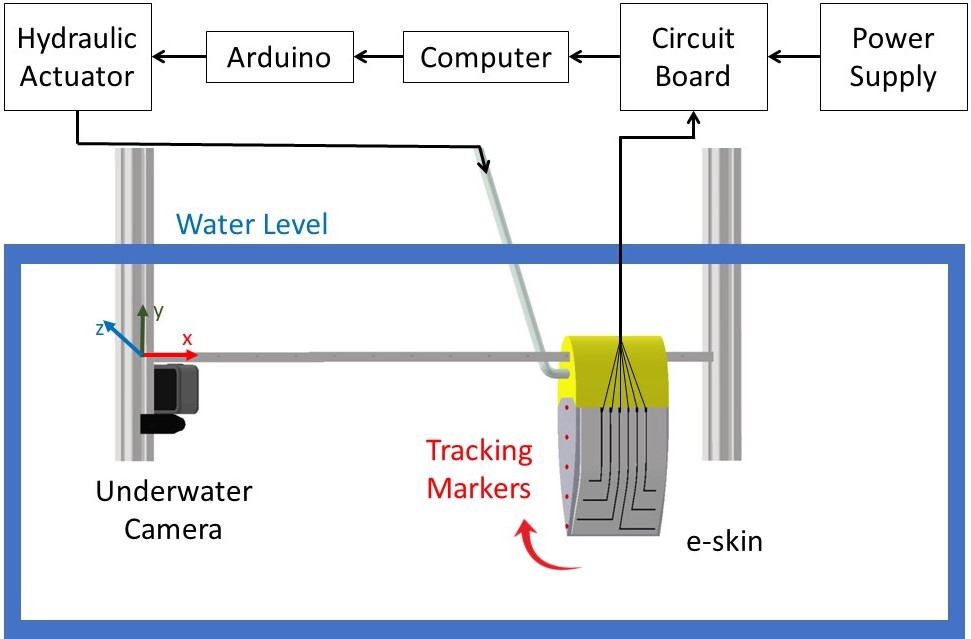}
    \caption{Schematic of the static testing set up. The pressurisation of the foil is controlled using a linear actuator connected to a syringe which supplies the pressure. An underwater camera records the motion of the foil for training and ground truth comparison.}
    \label{fig:test_set_up}
\end{figure}

Fig.~\ref{fig:test_set_up} shows the experimental setup for training and testing of the e-skin on the soft robotic foil. For the collection of training data, the foil is actuated for 10 cycles with 30 seconds of baseline data taken before and after actuation. The pressurisation of the foil is controlled using a linear actuator connected to a syringe which supplies the pressure. An underwater camera records the motion of the foil for training and ground truth comparison. The e-skin samples at 714\,Hz and the camera at 30\,Hz. The video is post-processed to track the five red markers and convert these to sets of five two-dimensional coordinates for each image. The coordinate sets and time-corresponding training data are then employed to train a Multilayer Perceptron (MLP) model which ultimately allows association of instantaneous capacitance readouts to foil shape, thus removing the need for visual feedback. 

We employ the MLP model because it is well-suited for handling structured data like the sensor readings from the e-skin, and it is able to capture complex non-linear relationships between the input sensor data and the output coordinates which would be expected from a more complex three-dimensional problem. Compared to recurrent or convolutional neural networks, the MLP is an effective, and more simplistic, choice for accurately estimating the foil shape from the sensor data as well as future 3D applications. The MLP model (Table~\ref{table: MLP}) has 1 input layer, 3 hidden layers, and 1 output layer. The input of this model is the 9 calibrated capacitance readouts in one frame from the e-skin. The output is a vector with a size of 10, indicating the coordinates of the five markers which can be used to calculate the camber.

\begin{table}[b]
\centering
\caption{TRAINING SPECIFICATIONs}
\label{table: MLP}
\begin{tabular}{cc}
\hline
 \multicolumn{2}{c}{MLP}\\
 \hline
Neurons per hidden layer   & 32, 128, 32             \\
Training Method & Mean Squared Error (MSE)\\ & loss function \\
Learning Rate & 0.0001 \\
Epochs & 300,000\\
Batch size & 256\\
Training Data & 24,000 frames\\
Training, Validating, Testing Ratio & 70/20/10\\
Minimum validation loss & 0.397 \\

\hline
 \multicolumn{2}{c}{PID Controller Gains}\\
  \hline
  
 $K_p$, $K_i$, $K_d$ & 50, 1, 1 \\

\end{tabular}
\end{table}


During training the foil leading edge is kept stationary, allowing for the camber line of the foil to be accurately described by 6 geometrical points. 
A spline is fitted to populate the points from the trailing edge to the start of the silicone and the chord line is calculated from the leading edge to the trailing edge, Fig. \ref{fig:camber explanation}; eqn.~\ref{eqn:camber} provides instantaneous estimate of the foil camber based on these two lines. 

\subsection{Underwater Soft Robotic Shape Control}

For a potential real world application the system needs to be able to reach a desired set point with different time scales and paths. For this work we tested a step function with 2\% camber increases every 5 seconds, and triangle and sinusoidal motion profiles. The motion profiles were tested at 20~s, 10~s, 5~s periods around a mean value of 4.25\% and amplitudes of 5\% and 2\%. Faster response times were not achievable due to the speed limitations of the linear actuator used to control the pressure. 

\begin{figure}[t!]
    \centering
    \includegraphics[width=8.0cm]{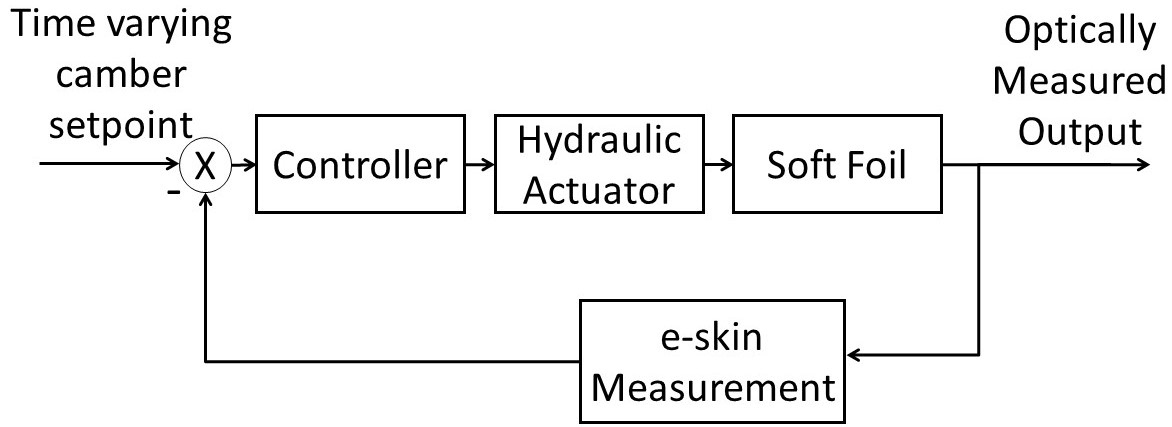}
    \caption{Block diagram outlining the closed loop control structure of the system. The system is comprised of an Arduino controller, a linear actuator driven syringe, and the soft robotic foil. The e-skin sensor provides feedback for the system.  }
    \label{fig:control Structure}
    \vspace{-0.4cm}
\end{figure}

The control loop is depicted in Fig. \ref{fig:control Structure}: relative capacitance measurements are read and the MLP model converts the signals into coordinates from which the camber is calculated. A PID controller compares the measured position to the set point and drives an input to the  hydraulic actuator. Parameters for the MLP and PID were determined through preliminary experiments, and are given in Table~\ref{table: MLP}.

\vspace{5mm}
\section{Results and Discussion}

\begin{figure}[b!]
    \centering
    \includegraphics[width=8.cm]{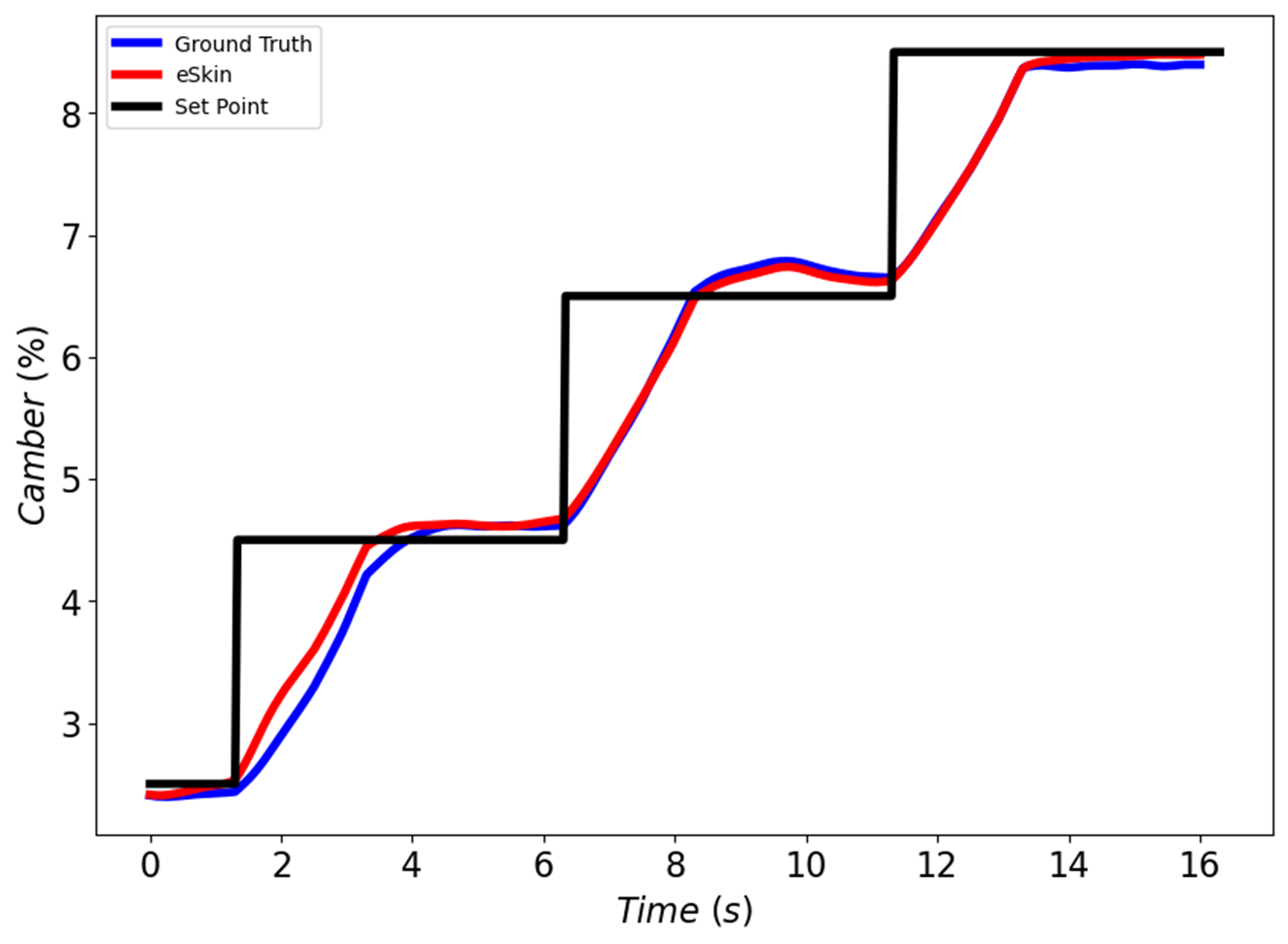}
    \caption{Plot of camber against time for a 2\% camber step increase every 5 seconds from 2.5\% to 8.5\% (equivalent to a tip amplitude change of 10~mm per step) with a rise time of 1.7~s. Plotted is the instantaneous foil camber based on the e-skin and the ground truth measurement from the camera point-tracking, averaged across 10 trials.  
    }
    \label{fig:Step Function}
\end{figure}

\begin{figure}[t!]
    \centering
    \includegraphics[width=8.5cm]{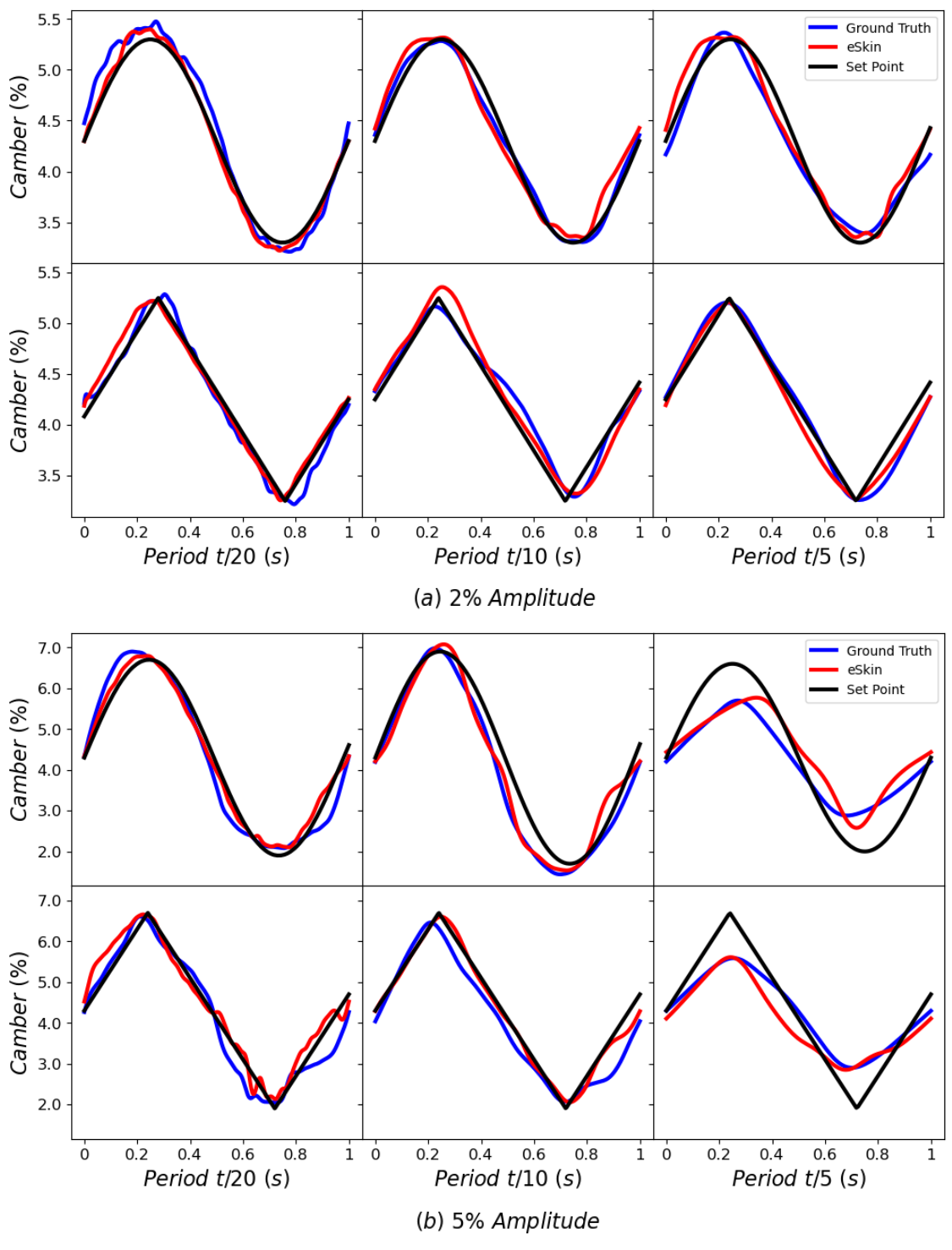}
    \caption{(a) Mean phase averaged camber plotted for six set point motion profiles. The top row correspond to Sinwave inputs. The bottom row corresponds to Triangle wave inputs. Each profile has a peak-to-peak amplitude of 2\% and a mean of 4.25\%. Each column corresponds to a period of 20~s, 10~s, 5~s from left to right respectively. Plotted is the estimated position of the robot based on the e-skin measurement and the ground truth position of the robot based on the camera point-tracking, averaged across 20 cycles. 
    (b) The same motion profiles are plotted, changing the peak-to-peak amplitude to 5\%. 
    }
    \label{fig:waveform results}
\end{figure}

Fig.~\ref{fig:Step Function} shows camber time history for a 2\% camber step increase every 5 seconds from 2.5\% to 8.5\%. This is equivalent of a tail tip position change of 10~mm per step. Plotted is the estimated position of the robot based on the e-skin measurement and the ground truth position of the robot based on the camera point-tracking, averaged across 10 trials. The rise time is 1.7~s. 

\begin{figure}[t!]
    \centering
    \includegraphics[width = 8.cm]{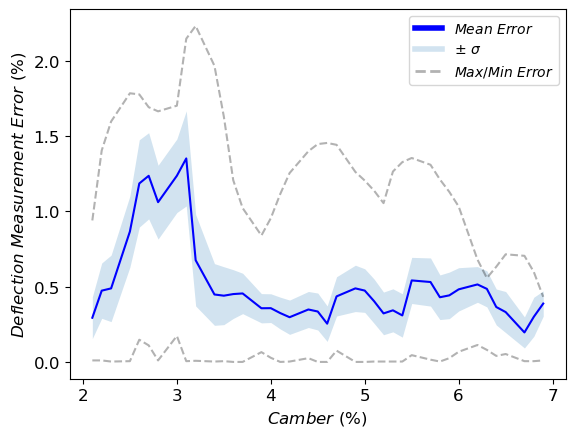}
    \caption{Comparison of the measurement error between the e-skin and the ground truth as a percentage of the foil length at different camber magnitudes. Plotted is the mean error, the first standard deviation, $\sigma$, of the error and the maximum and minimum error values measured. 
    }
    \label{fig:Sensor Error}
\end{figure}

\begin{figure}[t!]
    \centering
    \includegraphics[width = 8.cm]{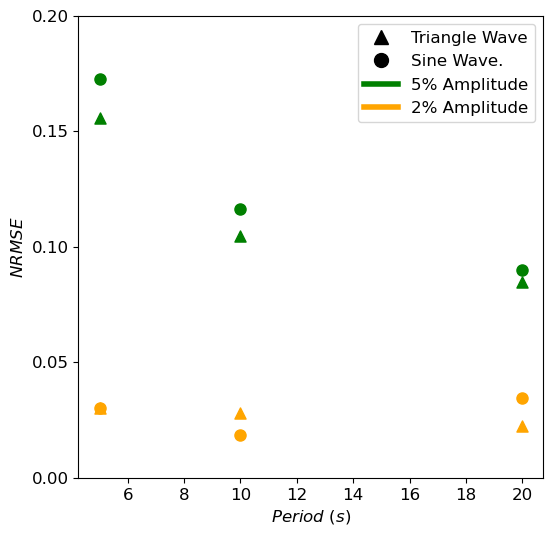}
    \caption{Comparing the error between set point and ground truth measurement for triangle and sinusoidal wave inputs at 2\% and 5\% camber variation. Plotted is the Root Mean Square Error normalised by the average signal magnitude: $RMSE/\Bar{y}$ (NRMSE). 
    }
    \label{fig:error summary}
\vspace{-0.3cm}
\end{figure}

Fig.~\ref{fig:waveform results}a shows the mean phase averaged camber plotted for six set point motion profiles. The top row corresponds to Sinwave inputs. The bottom row corresponds to Triangle wave inputs. Each profile has a peak-to-peak amplitude of 2\% and a mean of 4.25\% and each column corresponds to a period of 20~s, 10~s, 5~s from left to right respectively. Plotted is the estimated position of the robot based on the e-skin measurement and the ground truth position of the robot based on the camera point-tracking, averaged across 20 cycles. These results show excellent set point regulation for the 10~s period and even for the faster 5~s actuation routine. 

Fig.~\ref{fig:waveform results}b shows mean phase averaged camber plotted for the same six set point motion profiles, with a new peak-to-peak amplitude of 5\%, averaged across 20 cycles. 
Positional error observed for the 5~s period are mainly due to the slow response of the hydraulic actuator, as expected given the rise time measured in Fig.~\ref{fig:Step Function}. Discrepancies for the 20~s and 10~s cases for both the 2\% and 5\% test are attributed to high frequency signal noise.

Fig.~\ref{fig:Sensor Error} reports on the e-skin state estimation performance by comparing the measurement error between the e-skin and the ground truth as a percentage of the foil length at different camber magnitudes. The foil is calibrated based on the zero position so there is higher accuracy at this point. Plotted is the mean error, the first standard deviation, $\sigma$, of the error and the maximum and minimum error values measured. We demonstrate a maximum sensor error of less than 2.2\% and an average sensor error of 0.52\%. These errors are extremely small and sufficient given hydrodynamic considerations in unsteady natural flow. At relatively low camber values there is smaller overall deformation of the foil which leads to a worse signal-to-noise ratio and induces higher average error. 

Fig.~\ref{fig:error summary} assesses the closed-loop control performance by comparing the error between set point and ground truth measurement for triangle and sinusoidal wave inputs at 2\% and 5\% camber variation. Plotted is the Root Mean Square Error normalised by the average signal magnitude: $RMSE/\Bar{y}$ ~(NRMSE). NRMSE decreases with a longer period due to the speed of actuation capabilities. NRMSE also decreases with a smaller desired amplitude due to the slower operation required and more linear response of the system. There is little difference in error between the triangle and sinusoidal wave setpoints. 

The results presented give confidence in the robustness and accuracy of the underwater e-skin for the purpose of control of the flexible wing. Further extension of the operating envelope of the system would be feasible through revision of the actuator and optimization of the control strategy. 

\vspace{5mm}
\section{Conclusion and Robotic Implications}

We have demonstrated the ability to perform real time shape estimation and control of a soft aquatic propulsor using an integrated e-skin, without the need for external sensing capabilities. We demonstrate measurements of foil camber with a maximum sensor error of less than 2.2\% and an average sensor error of 0.52\% which is comparable to that produced by an external optical system. The e-skin captures the high non-linearity of the soft robotic foil with a NRMSE of 0.11 and is not limited to small deformations. This method of underwater shape estimation is well suited for progression to more complex systems as it is agnostic to the shape of the system, making it suitable for more complex shapes and spatial deformations. Expansion to three-dimensional state-estimation is exclusively limited by the need to undertake training underwater, thus requiring adequate underwater motion tracking technology. Since this novel approach does not rely on external sensors, it paves the way for future real time state estimation and closed-loop control of aquatic soft robots for propulsive efficiency optimisation and disturbance rejection.

\bibliographystyle{unsrt}
\bibliography{References}

\end{document}